\documentclass[10pt,twocolumn,letterpaper]{article}
\usepackage[pagenumbers]{cvpr}

\usepackage{graphicx}
\graphicspath{{Figures/}}

\usepackage{booktabs}
\usepackage{amsfonts}
\usepackage{amsmath}
\usepackage{amssymb}
\usepackage{multirow}
\usepackage{xcolor}
\usepackage{rotating}
\usepackage{array}

\def\paperID{*****}
\def\confName{CVPR}
\def\confYear{2026}

\definecolor{cvprblue}{rgb}{0.21,0.49,0.74}
\usepackage[pagebackref,breaklinks,colorlinks,allcolors=cvprblue]{hyperref}

\title{C$^2$Path: Class-Conditional Pathway Decoupling for Vision-Language\\Incremental Object Detection}

\author{
 Lecheng Xu$^{1}$, \;Feifei Shao$^{1}$\thanks{Feifei Shao is the corresponding author.}\;, Ouyangzi Ye$^1$, \;Zhen Wang$^{2}$, \;Lin Li$^{2}$, \\
 Kexin Li$^{3}$, \;Zhao Wang$^1$, \;Changqin Huang$^1$ \\
  \small $^1$ Zhejiang University\; 
  \small $^2$ The Hong Kong University of Science and Technology\;
  \small $^3$ Zhejiang Tobacco Monopoly Administration\; \\
  \small \texttt{\{lechx, sff, yeouyangzi, zhao\_wang, cqhuang\}@zju.edu.cn, } \\
  \small \texttt{\{zhenwang, lllidy\}@ust.hk, likexin@zjyc.cn} \\
  }

\begin{document}

\maketitle


\begin{abstract}
Incremental Object Detection (IOD) aims to enable detectors to continuously learn novel categories while preserving previously acquired knowledge. However, existing methods suffer from two forms of \textbf{class knowledge coupling}: class boundary erosion induced by shared parameter updates and class representation entanglement arising from mixed feature encoding. We argue that effective incremental learning requires class-specific computational pathways that enable isolated parameter updates and separated class-wise injection. To this end, we propose \textbf{C$^2$Path}, a class-conditional pathway decoupling framework for vision-language incremental object detection that leverages token-level class cues to establish dedicated and updatable computational pathways for different categories. Specifically, C$^2$Path introduces a category expert library and a class-conditional decoupling module. The expert library consists of learnable low-rank computational nodes that capture category-specific knowledge, while the decoupling module generates class-aware routing signals to dynamically compose \textit{ClassLoRA} adapters from these experts, thereby forming class-specific computational pathways for isolated updates and separated injection across categories. Extensive experiments on COCO 2017 under multiple incremental learning settings demonstrate that C$^2$Path consistently outperforms state-of-the-art methods, providing an effective and scalable solution for continual category expansion in vision-language detectors.
\end{abstract}

\section{Introduction}


Traditional object detection models~\cite{zou2023object, zhao2019object} are typically trained on a fixed set of predefined categories and assume a static data distribution. However, real-world intelligent systems must operate in dynamic environments, where they continuously encounter new categories and evolving data streams \cite{parisi2018continual}. Updating detectors with novel categories often leads to catastrophic forgetting~\cite{kirkpatrick2017overcoming, kemker2018measuring}, causing significant degradation in previously learned categories. To overcome this limitation, Incremental Object Detection (IOD) has emerged as a fundamental paradigm that enables detectors to acquire new knowledge continuously while retaining previously learned capabilities.

The evolution of IOD reflects a gradual transition from preserving feature-level stability to maintaining semantic-level consistency in increasingly open and multimodal environments. Early studies primarily focused on mitigating catastrophic forgetting in closed-set detectors by preserving learned representations, employing functional distillation for two-stage architectures \cite{shmelkov2017incremental,peng2020faster,li2017learning} and intermediate feature regularization for anchor-free frameworks \cite{peng2021sid,feng2022overcoming}. As incremental scenarios expanded beyond fixed label spaces, subsequent works explored open-world discovery and meta-learning paradigms to improve adaptability to emerging categories \cite{joseph2021towards,joseph2021incremental}. More recently, Vision-Language Models (VLMs) \cite{radford2021learning} have further reshaped IOD by replacing fixed classification heads with prompt-driven cross-modal alignment mechanisms \cite{deng2024zero,kim2024vlm}.

\begin{figure}[t]
    \centering
    \includegraphics[width=1\columnwidth]{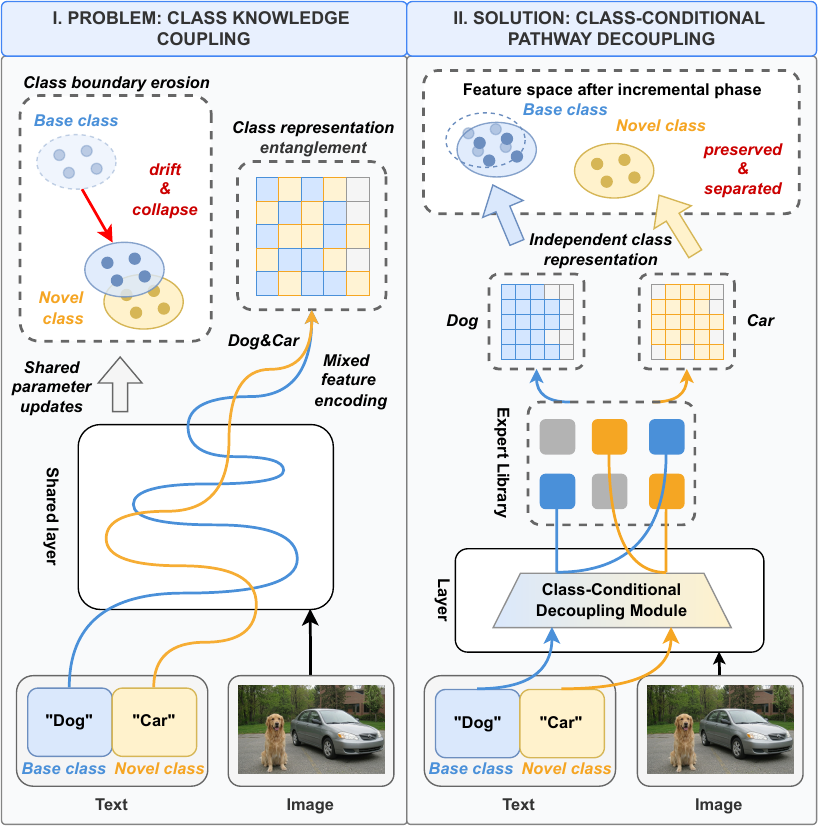}
    \caption{Illustration of class knowledge coupling problem and class-conditional pathway decoupling solution. Existing methods suffer from class boundary erosion and representation entanglement. C$^2$Path dynamically selects category-specific experts through class-conditional routing to construct independent pathways, enabling novel-category adaptation while preserving existing class knowledge.}\label{fig:drift_interference}
\end{figure}

Taking the vision-language IOD method GCD \cite{wang2025gcd} as an example, it introduces global semantic alignment through external distillation to maintain semantic consistency across incremental phases. However, GCD still relies on shared computational pathways, where both parameter adaptation and feature encoding are performed in a category-agnostic manner. As illustrated in Figure~\ref{fig:drift_interference}, such category-agnostic design lacks the granularity required for class-specific learning, causing \textbf{class knowledge coupling} at both the parameter and representation levels. At the parameter level, shared parameter updates during novel-category learning inevitably modify previously acquired knowledge, leading to class boundary erosion. At the representation level, jointly encoding multiple category representations mixes class-specific semantics, causing representation entanglement and degrading class discrimination.

To address the above limitation, we propose \textbf{C$^2$Path}, a class-conditional pathway decoupling framework for vision-language incremental object detection. Inspired by mixture-of-experts architectures, C$^2$Path leverages token-level class cues to establish dedicated and updatable computational pathways for different categories, enabling isolated parameter updates and separated class-wise injection. Specifically, C$^2$Path parameterizes each class-specific pathway using dynamically synthesized \textit{ClassLoRA} adapters. Unlike conventional LoRA \cite{hu2022lora}, which applies a single static low-rank update shared across all categories, C$^2$Path generates category-specific adaptation weights conditioned on token-level class cues at both training and inference time. To achieve this, C$^2$Path introduces a category expert library and a class-conditional decoupling module. The expert library contains learnable low-rank computational nodes that capture category-specific knowledge, while the decoupling module generates class-aware routing signals to dynamically compose \textit{ClassLoRA} adapters from these experts. The synthesized \textit{ClassLoRA} are subsequently integrated into the detection backbone, establishing class-specific computational pathways to facilitate continual category expansion.

The contributions of this paper are summarized as follows:
\begin{itemize}
\item We propose C$^2$Path, a novel class-conditional pathway decoupling framework for vision-language incremental object detection that mitigates class knowledge coupling by constructing dedicated and dynamically updatable computational pathways for different categories.

\item C$^2$Path introduces two key components: a category expert library and a class-conditional decoupling module. The former captures diverse category-specific knowledge through low-rank expert nodes, while the latter generates semantic-aware routing signals to dynamically compose \textit{ClassLoRA} adapters, enabling localized parameter updates and separated class-wise injection.

\item Extensive experiments on MS COCO 2017 demonstrate that C$^2$Path achieves state-of-the-art performance across multiple incremental learning settings, including 40+40 and 70+10 category splits.
\end{itemize}

\section{Related Work}

\subsection{Incremental Object Detection}

Early IOD methods build on closed-set detectors and rely on knowledge
distillation \citep{li2017learning} to preserve base-category knowledge,
including functional distillation for two-stage architectures
\citep{shmelkov2017incremental,peng2020faster}, selective distillation for
anchor-free frameworks \citep{peng2021sid,feng2022overcoming}, and
extensions to open-world discovery and meta-learning
\citep{joseph2021towards,joseph2021incremental}. With Transformers,
query-based designs emerged: OW-DETR \citep{gupta2022ow} and CL-DETR
\citep{liu2023continual} tailor novelty identification and distillation to
DETR, DMD \citep{kang2023alleviating} adopts decoupled multi-level
distillation, and SDDGR \citep{kim2024sddgr} synthesizes pseudo-exemplars
via diffusion-based generative replay.

More recently, Vision-Language Detectors (VLDs) such as GLIP
\citep{li2022glip} and Grounding DINO \citep{liu2024grounding} redefine
detection as language-guided query matching \citep{radford2021learning},
inspiring a new line of IOD research. VLM-PL \citep{kim2024vlm} generates
pseudo-labels for previously seen categories, TALIR \citep{zhang2024learning}
aligns the language embedding space with incremental objectives, ZiRa
\citep{deng2024zero} designs a dual-branch structure with a zero-interference
loss, MR-GDINO \citep{dong2024mr} maintains task-specific memories, and GCD
\citep{wang2025gcd} achieves state-of-the-art performance via global
alignment and correspondence distillation, which we retain as our
continual-learning baseline.

Despite architectural differences, both lines share a common limitation: knowledge is isolated at most at the \emph{task} level, with all categories flowing through shared parameters---closed-set methods impose uniform updates that allow novel gradients to overwrite base-class boundaries, while VLD-based methods (even task-specific memories such as MR-GDINO) do not condition fusion-layer updates on category semantics. This leaves category-level drift and token-level interference unresolved, which we address by conditioning parameter synthesis on individual category tokens.

\subsection{Parameter-Efficient Continual Learning}

Low-Rank Adaptation (LoRA) \citep{hu2022lora} has become the standard for
parameter-efficient fine-tuning, with growing extensions to continual
learning. Regularization-based approaches impose orthogonality or
interference-free constraints on sequential low-rank updates
\citep{wang2023olora,he2025cl,liang2024inflora}, while composition-based
approaches route among task-specific or mixture-of-LoRA experts
\citep{yu2024boosting}, self-expand with new adapters \citep{wang2025self},
or decouple base and novel updates into separate low-rank components
\citep{wu2025sd}. Our work is also related to hypernetworks
\citep{ha2017hypernetworks,von2019continual}, which generate weights from
conditioning inputs, and to prompt-based continual learning
\citep{wang2022l2p}; notably, CODA-Prompt \citep{smith2023coda} composes
shared prompt components via attention-weighted combinations, sharing the
compositional spirit of our basis-tensor design. However, these methods
condition adaptation on task identities, global instance features, or
query-key matching in the prompt space, and thus do not synthesize
fusion-layer updates from individual class-name tokens within a multi-class
detection prompt, nor exploit the semantics carried by category names in VLD
prompts. In contrast, \textit{ClassLoRA} synthesizes LoRA weights on-the-fly from
individual category-token embeddings and injects them into cross-modality
fusion layers, achieving category-token-conditioned soft decoupling rather
than task-level or prompt-level composition.

\section{Methodology}
\begin{figure*}[t]
    \centering
    \includegraphics[width=1.0\textwidth]{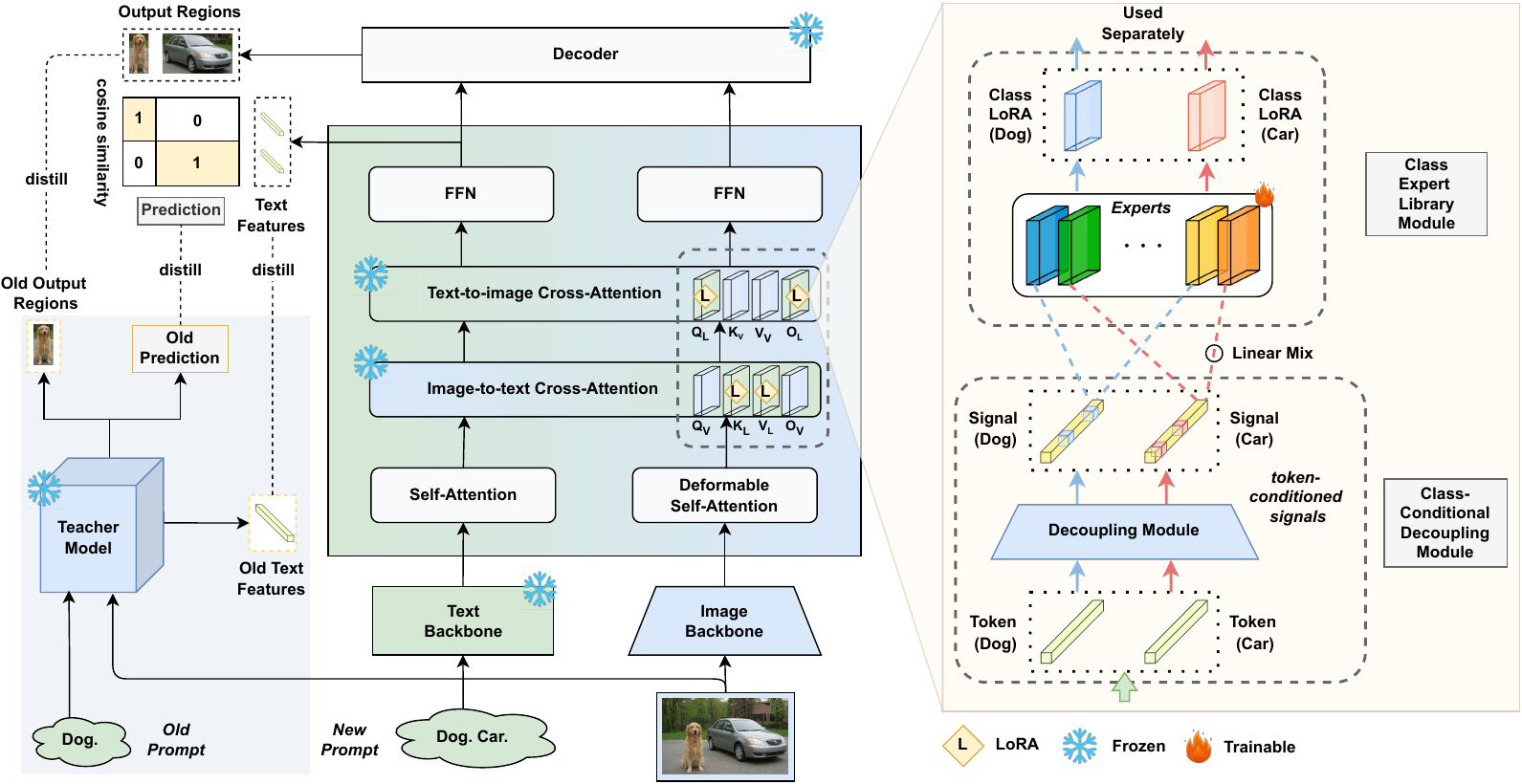}
    \caption{Overview of the C$^2$Path framework. Left: the overall incremental learning framework with multi-level knowledge distillation. Right: the proposed category-conditioned parameter synthesis mechanism, where the Class-Conditional Decoupling Module (\textbf{CCDM}) transforms intermediate text features into token-conditioned activation signals, which dynamically combine low-rank experts in the Class Expert Library (\textbf{CEL}) to synthesize token-specific \textit{ClassLoRA} instances.}\label{fig:overview}
\end{figure*}
\subsection{Preliminaries}

\textbf{Problem Setting.} We formulate Incremental Object Detection (IOD) as a two-stage learning process consisting of a base phase and an incremental phase. During the base phase, the model is optimized on the base dataset $\mathcal{D}_{\mathrm{base}}$ covering the base category set $\mathcal{C}_{\mathrm{base}}$. During the incremental phase, the model is exposed to the novel dataset $\mathcal{D}_{\mathrm{novel}}$, where only objects belonging to the novel category set $\mathcal{C}_{\mathrm{novel}}$ are annotated. Although the label spaces are disjoint ($\mathcal{C}_{\mathrm{base}} \cap \mathcal{C}_{\mathrm{novel}} = \emptyset$), the underlying images in the two datasets may overlap, meaning that previously learned categories can still appear in $\mathcal{D}_{\mathrm{novel}}$ without annotations. Our objective is to develop a robust vision-language detector that preserves performance on $\mathcal{C}_{\mathrm{base}}$ while effectively learning $\mathcal{C}_{\mathrm{novel}}$, ultimately achieving high localization and classification accuracy over the combined category set $\mathcal{C}_{\mathrm{base}} \cup \mathcal{C}_{\mathrm{novel}}$.

\textbf{Vision-Language Detection Model}. In this work, we use Grounding DINO \cite{liu2024grounding} as our detection model, which reformulates object detection as a language-guided query-matching task. The
architecture comprises a vision backbone, a text backbone, a cross-modality
encoder, and a cross-modality decoder. The cross-modality encoder is
instrumental for feature alignment, utilizing Bi-Directional Cross-Modality
Layers to perform deep fusion via bidirectional multi-head attention.
Specifically, let $F_V \in \mathbb{R}^{N_v \times d}$ and
$F_L \in \mathbb{R}^{N_l \times d}$ denote the visual and language features,
respectively. The bidirectional attention is computed as:
\begin{equation}
    \text{Attn}(F_V, F_L) = \text{Softmax}\left(\frac{Q_V K_L^{\top}}{\sqrt{d_k}}\right) V_L,
\end{equation}
\begin{equation}
    \text{Attn}(F_L, F_V) = \text{Softmax}\left(\frac{Q_L K_V^{\top}}{\sqrt{d_k}}\right) V_V,
\end{equation}
where $Q$, $K$, and $V$ denote the query, key, and value projections, the
subscripts indicate the source modality, and $d_k$ is the dimensionality of the
key vectors. The cross-modality decoder utilizes these fused features to
generate object queries.


\subsection{Technical Details of C$^2$Path}

To address the class knowledge coupling issue illustrated in Figure~\ref{fig:drift_interference}, we propose a class-conditional pathway decoupling framework, termed C$^2$Path, for vision-language incremental object detection. Its pipeline proceeds as follows: (1) intermediate text features from each fusion layer are normalized and fed into the Class-Conditional Decoupling Module to generate sparse token-conditioned routing signals; (2) these signals activate and combine low-rank experts from the Category Expert Library to synthesize token-specific adaptation modules, \textit{i.e., ClassLoRA}, which define class-specific parameter transformations; (3) The synthesized \textit{ClassLoRA} adapters are injected into the language-side projections of the fusion layers. Through semantic-conditioned expert composition and adaptive parameter generation, C$^2$Path realizes class-specific computational pathways, enabling localized updates, reducing inter-class interference, and alleviating representation entanglement during open-ended category expansion.


\subsubsection{Category-Conditional Decoupling Module (CCDM).}

To avoid the feature intermixing induced by shared weight updates, we condition token-specific adapter generation on the
intermediate text features within each fusion layer. Concretely, at each
Bi-Directional Cross-Modality Layer, let $f_t \in \mathbb{R}^{d}$ denote the
$t$-th token state of the text features $F_L$. The decoupling module first
applies $\ell_2$ normalization:
\begin{equation}
    \tilde{f}_{t} = \frac{f_{t}}{\|f_{t}\|_2 + \epsilon},
\end{equation}
where $\epsilon$ ensures numerical stability. This removes token-wise magnitude
variation, so that the subsequent routing is driven by the semantic orientation
of $f_t$ rather than its norm. To capture semantic non-linearities and
selectively route tokens to relevant adaptation subspaces, we transform the
normalized representation with a two-layer gated projection, followed by a
constrained sparse selection:
\begin{align}
    h_t &= \mathbf{W}_2\,\text{GELU}\left(\mathbf{W}_1 \tilde{f}_t\right)
    \in \mathbb{R}^{M},\\
    \pi_t &= \text{Softmax}\left(\text{TopK}\left(h_t,k\right)\right)
    \in \mathbb{R}^{M},
\end{align}
where $M$ is the number of routing channels and
$\text{TopK}(\cdot,k)$ retains the $k$ largest entries while setting the
remainder to $-\infty$, yielding a sparse probability distribution after
Softmax. We denote by
$\mathcal{S}_t \subset \{1,\dots,M\}$ the index set of the retained entries.
The resulting sparse activation signal $\pi_t$ serves as a
category-conditioned routing distribution that determines the subsequent
adaptation pathway for each token. Tokens with similar semantics naturally
produce similar routing distributions, whereas semantically distinct
categories activate different expert sets $\mathcal{S}_t$. Consequently, the routing mechanism partitions the linguistic feature
space, inducing category-dependent adaptation patterns without introducing
explicit class-specific parameters, thereby reducing feature interference
caused by shared adaptation weights.

\subsubsection{Category Expert Library.}
It can be viewed as a compact dictionary of
adaptation primitives, where each expert captures a distinct mode of
cross-modal alignment. Given the routing distribution $\pi_t$, a
token-specific \emph{ClassLoRA} instance is synthesized as a sparse
combination of shared basis components and subsequently injected into the
cross-modal fusion layers through a category-wise separated injection
strategy. Unlike conventional adaptation schemes that employ shared
adaptation parameters across all categories, the proposed strategy assigns
independently synthesized \textit{ClassLoRA} instances to different semantic tokens,
thereby preventing heterogeneous category knowledge from being injected
through a common adaptation pathway.

To avoid a linear increase in parameter overhead, the library is
parameterized by learnable basis tensors
$\mathcal{B}_A \in \mathbb{R}^{M \times r \times d_{\text{in}}}$ and
$\mathcal{B}_B \in \mathbb{R}^{M \times d_{\text{out}} \times r}$, where each
pair of frontal slices
$(\mathcal{B}_A)_j$ and $(\mathcal{B}_B)_j$
constitutes a low-rank expert and $r$ denotes the adaptation rank. For a
target projection layer, the token-specific adaptation matrices
$\mathbf{A}_t \in \mathbb{R}^{r \times d_{\text{in}}}$ and
$\mathbf{B}_t \in \mathbb{R}^{d_{\text{out}} \times r}$ are synthesized as
\begin{equation}
    \mathbf{A}_{t}
    =
    \sum_{j\in\mathcal{S}_t}
    \pi_{t,j}(\mathcal{B}_A)_j,
    \qquad
    \mathbf{B}_{t}
    =
    \sum_{j\in\mathcal{S}_t}
    \pi_{t,j}(\mathcal{B}_B)_j,
\end{equation}
where $\mathcal{S}_t$ denotes the selected expert set for token $t$.
The resulting pair $\{\mathbf{A}_t,\mathbf{B}_t\}$ forms a
\emph{ClassLoRA} instance assigned exclusively to token $t$, defining the
sparse, differentiable mapping
\begin{equation}
    \mathcal{G}:\pi_t\mapsto
    \{\mathbf{A}_t,\mathbf{B}_t\}.
\end{equation}

The resulting \emph{ClassLoRA} instance is introduced through the proposed
category-wise separated injection mechanism within the
Bi-Directional Cross-Modality Layers, where cross-modal attention establishes
the alignment between linguistic semantics and visual evidence in Grounding
DINO \cite{liu2024grounding}. Specifically, the synthesized \textit{ClassLoRA}
parameters are applied only to the language-side projections involved in
cross-modal interaction, enabling category-dependent semantic modulation
while keeping the visual-side pathway unchanged. Restricting adaptation to
these fusion layers concentrates model capacity on cross-modal alignment
while avoiding unnecessary modifications to unimodal representations.
Furthermore, only the language-side projections are adapted, whereas the
visual projections remain frozen. Since incremental learning primarily
introduces new category semantics through language representations, while
low-level visual patterns (\textit{e.g.,} edges, textures, and shapes) are
less sensitive to category-specific variations, language-side adaptation
effectively guides cross-modal interaction while preserving stable visual
representations.

Given the synthesized matrices, the token-wise modulation of a language-side
projection is defined as
\begin{equation}
    \left[\Phi^{*}(F_L)\right]_{t}=\gamma\cdot\mathbf{B}^{*}_{t}\mathbf{A}^{*}_{t}f_{t},
\end{equation}
where $f_t$ denotes the feature representation of the $t$-th token in the
language feature sequence $F_L$, $\gamma$ is a scaling factor, and the
superscript $*$ indicates the projection-specific modulation with the
corresponding synthesized \textit{ClassLoRA} matrices adapted to each target
projection layer.

For $\text{Attn}(F_V, F_L)$, the language-side keys and values are modulated
before attention, producing the updated visual representation $H_V$:
\begin{equation}
\begin{aligned}
    K_L' &= F_L W_{K_L} + \Phi_{K_L}(F_L), \\
    V_L' &= F_L W_{V_L} + \Phi_{V_L}(F_L), \\
    H_V &=
    \mathrm{Softmax}
    \left(
    \frac{Q_V{K_L'}^{\top}}{\sqrt{d_k}}
    \right)
    V_L'W_{O_V}.
\end{aligned}
\end{equation}

For $\text{Attn}(F_L, F_V)$, the language-side query and output
projections are modulated, while the visual keys and values remain
unadapted, producing the adapted language representation $H_L$:
\begin{equation}
\begin{aligned}
    Q_L' &= F_L W_{Q_L} + \Phi_{Q_L}(F_L), \\
    Z_L &=
    \mathrm{Softmax}
    \left(
    \frac{Q_L'K_V^{\top}}{\sqrt{d_k}}
    \right)
    V_V, \\
    H_L &= Z_L W_{O_L}+\Phi_{O_L}(Z_L),
\end{aligned}
\end{equation}
where $W_*$ denotes the frozen projection matrices in the cross-modal
attention layers.

By synthesizing \textit{ClassLoRA} instances from a shared expert library, the
proposed design restricts adaptation generation to a structured basis space,
providing a compact and reusable representation for category-specific
adaptation. Furthermore, token-specific expert assignment encourages
different semantic groups to activate distinct subsets of adaptation bases,
thereby disentangling parameter updates and alleviating cross-category
interference during incremental learning. Through category-wise separated
injection, the proposed framework enables incremental semantic expansion by
allowing novel category knowledge to be incorporated through dedicated
adaptation pathways while maintaining previously learned representations.

\subsubsection{Model Training.}


Following the training paradigm of GCD \cite{wang2025gcd}, we train C$^2$Path using a joint optimization objective that combines the standard detection losses with a multi-modal distillation loss. Specifically, the overall training objective is defined as:
\begin{equation}
\mathcal{L}_{total} =
\mathcal{L}_{cls}
+\mathcal{L}_{reg}
+\mathcal{L}_{distill},
\end{equation}
where $\mathcal{L}_{cls}$,  $\mathcal{L}_{reg}$, and $\mathcal{L}_{distill}$ denote the classification, regression and multi-modal distillation loss, respectively.

\textbf{Detection Losses.}
Following standard DETR-style object detection training \cite{carion2020end}, we employ classification and box regression losses, where $\mathcal{L}_{cls}$ optimizes classification confidence and $\mathcal{L}_{reg}$ optimizes localization accuracy through L1 and GIoU losses. For each matched object prediction, the losses are defined as:
\begin{equation}
\begin{aligned}
\mathcal{L}_{cls} &= 
\mathrm{Focal}(\hat{s}, s),\\
\mathcal{L}_{reg} &=
\lambda_{\mathrm{L1}}\cdot\mathcal{L}_{\mathrm{L1}}(\hat{b}, b)
+\lambda_{\mathrm{GIoU}}\cdot\mathcal{L}_{\mathrm{GIoU}}(\hat{b}, b),
\end{aligned}
\end{equation}
where $\hat{s}$ and $\hat{b}$ denote the predicted classification score and bounding box, while $s$ and $b$ represent the ground-truth label and box. $\lambda_{\text{L1}}$ and $\lambda_{\text{GIoU}}$ denote hyperparameters.

\textbf{Distillation Losses.} Following GCD \cite{wang2025gcd}, the distillation objective $\mathcal{L}_{distill}$ is formulated as:
\begin{equation}
\mathcal{L}_{distill} =  \mathcal{L}_{\text{CRD}} + \mathcal{L}_{\text{CTD}}.
\end{equation}
Specifically, $\mathcal{L}_{distill}$ consists of Correspondence Response Distillation (CRD) and Correspondence Topology Distillation (CTD) losses. The former is designed to distill alignment logits via KL divergence, while the latter is responsible for enforcing topological consistency by matching pair-wise prototype distance matrices of the student and teacher. CRD loss is computed as:
\begin{equation}
\mathcal{L}_{\text{CRD}} = \sum_{i=1}^{N} \alpha_i \cdot \text{KL}(P_i^{\text{old}} \| P_i),
\end{equation}
where $P_i^{old}$ and $P_i$ denote the softened logit distributions of teacher model and current model, respectively. $\alpha_i$ is the weights of each prediction. Regression outputs are distilled analogously and aggregated over $K$ layers for the total CRD loss.
CTD loss is computed as:
\begin{equation}
\mathcal{L}_{\text{CTD}} = \lambda_1\cdot\|R-R^{\text{old}}\|_2 + \lambda_2\cdot\|\hat{R}-\hat{R}^{\text{old}}\|_2,
\end{equation}
where $R$ and $\hat{R}$ are pairwise Euclidean distance matrices of class-wise object and text prototypes, respectively. $\lambda_1$ and $\lambda_2$ denote the corresponding hyperparameters.

\section{Experiments}

\subsection{Experimental Settings}

\paragraph{Datasets, Protocols, and Metrics.}
We evaluate C$^2$Path on MS COCO 2017 \cite{lin2014microsoft}, which contains 118k training images, 5k validation images, and 80 object categories, following the two standard two-phase IOD protocols introduced in GCD \cite{wang2025gcd}. Specifically, the 40+40 setting partitions categories 1--40 as base categories and 41--80 as novel categories, while the 70+10 setting uses 70 base categories followed by 10 novel categories. Both settings strictly follow the annotation protocol and exemplar-free constraint described in \S3.1. After incremental training, all models are evaluated on the complete val2017 set over all 80 categories using standard COCO metrics, including AP, $\text{AP}{50}$, $\text{AP}{75}$, and scale-specific $\text{AP}_{S/M/L}$.

\paragraph{Implementation Details.} Following the setup of Grounding DINO (Swin-T) \cite{liu2024grounding}, we adopt AdamW with an initial learning rate of $10^{-4}$ and a multi-step decay schedule over 12 epochs, using a total batch size of 8. During the base phase, the model is fully trained on base categories without introducing any C$^2$Path modules. During incremental phases, we jointly optimize the neck, detection head, Category Expert Library, Category-Conditional Decoupling Module, and textual self-attention blocks in the encoder. The visual backbone and visual self-attention blocks are fine-tuned with a $0.1\times$ learning-rate multiplier, while the text backbone and decoder remain frozen. For the low-rank experts, we set the rank $r=16$ and scaling factor $\gamma=32$. The decoupling module adopts a hidden dimension of 128, with $M=64$ experts and TopK sparsity $k=8$. Following GCD \cite{wang2025gcd}, the distillation coefficients are set to $\lambda_1=3$ and $\lambda_2=5$, while the regression loss weights are set to $\lambda_{\mathrm{L1}}=5$ and $\lambda_{\mathrm{GIoU}}=2$.

\begin{table*}[t]
\centering
\small
\setlength{\tabcolsep}{5pt}
\begin{tabular}{@{}c|lcccccccc@{}}
\toprule
Setting & Method & Detector & $AP$ & $AP_{50}$ & $AP_{75}$ & $AP_{S}$ & $AP_{M}$ & $AP_{L}$ \\ \midrule
\multirow{8}{*}{40+40} & CL-DETR \cite{liu2023continual} & Deformable DETR & 42.0 & 60.1 & 45.9 & 24.0 & 45.3 & 55.6 \\
 & DMD \cite{kang2023alleviating} & Deformable DETR & 39.8 & - & - & - & - & - \\
 & SDDGR \cite{kim2024sddgr} & Deformable DETR & 43.0 & 62.1 & 47.1 & 24.9 & 46.9 & 57.0 \\ \cmidrule{2-9}
 & TALIR \cite{zhang2024learning} & GLIP$^{*}$ & 40.4 & 57.4 & 43.9 & 23.3 & 44.7 & 54.5 \\
 & LwF \cite{li2017learning} & Grounding DINO$^{*}$ & 21.8 & 29.4 & 23.7 & 12.4 & 23.0 & 30.4 \\
 & ERD \cite{feng2022overcoming} & Grounding DINO$^{*}$ & 29.7 & 41.8 & 32.1 & 19.0 & 33.1 & 38.8 \\
 & GCD \cite{wang2025gcd} & Grounding DINO$^{*}$ & 45.7 & 62.9 & 49.7 & 28.4 & 49.3 & 60.0 \\
 & \textbf{C$^2$Path (Ours)} & Grounding DINO$^{*}$ & \textbf{46.5} & \textbf{63.2} & \textbf{50.7} & \textbf{29.8} & \textbf{49.8} & \textbf{60.6} \\
\midrule\multirow{8}{*}{70+10} & CL-DETR \cite{liu2023continual} & Deformable DETR & 40.4 & 58.0 & 43.9 & 23.8 & 43.6 & 53.5 \\
 & DMD \cite{kang2023alleviating} & Deformable DETR & 37.6 & - & - & - & - & - \\
 & SDDGR \cite{kim2024sddgr} & Deformable DETR & 40.9 & 59.5 & 44.8 & 23.9 & 44.7 & 54.0 \\ \cmidrule{2-9}
 & TALIR \cite{zhang2024learning} & GLIP$^{*}$ & 42.9 & 59.2 & 45.2 & 24.3 & 45.1 & 54.1 \\
 & LwF \cite{li2017learning} & Grounding DINO$^{*}$ & 11.4 & 16.6 & 12.1 & 7.7 & 13.6 & 18.0 \\
 & ERD \cite{feng2022overcoming} & Grounding DINO$^{*}$ & 39.0 & 53.6 & 42.1 & 24.7 & 42.2 & 53.2 \\
 & GCD \cite{wang2025gcd} & Grounding DINO$^{*}$ & 46.7 & 63.9 & 50.8 & 29.7 & 49.9 & 61.6 \\
 & \textbf{C$^2$Path (Ours)} & Grounding DINO$^{*}$ & \textbf{48.7} & \textbf{65.6} & \textbf{53.1} & \textbf{31.1} & \textbf{52.3} & \textbf{63.7} \\ \bottomrule
\end{tabular}
\caption{Comparison with state-of-the-art methods on MS COCO 2017. $^{*}$denotes
vision-language detectors.}
\label{tab:sota_comparison}
\end{table*}


\subsection{Comparison with State-of-the-Art Methods}

We compares C$^2$Path with state-of-the-art incremental object detection methods under two settings: 40+40 and 70+10. The former learns 40 base classes followed by 40 novel classes in an exemplar-free incremental phase, while the latter learns 70 base classes followed by 10 novel classes. In both settings, the final performance is evaluated over all 80 classes.

\textbf{40+40 Setting.} As reported in Table~\ref{tab:sota_comparison}, C$^2$Path achieves 46.5 AP, outperforming GCD by 0.8 AP under the same detector architecture and training protocol. It also consistently improves detection across different object scales, achieving 29.8, 49.8, and 60.6 AP for small, medium, and large objects, respectively. These results demonstrate that class-specific computational pathway decoupling enables effective knowledge adaptation while preserving learned representations.

\begin{figure}[t]

    \centering

    \includegraphics[width=\linewidth]{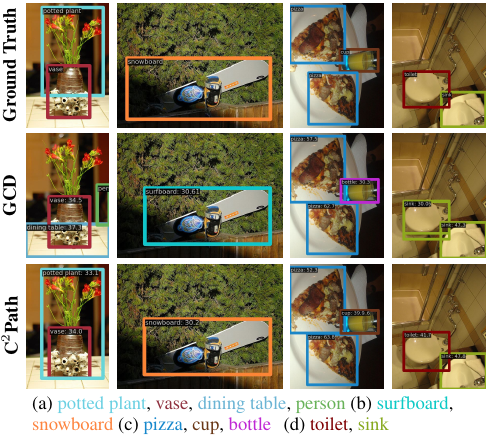}

    \caption{The labels (a--d) show enlarged views of the object categories predicted in the four columns from left to right.}

    \label{fig:qualitative}

\end{figure}

\textbf{70+10 Setting.} As reported in Table~\ref{tab:sota_comparison}, C$^2$Path further enlarges the performance gap, achieving 48.7 AP and outperforming GCD by 2.0 AP. This substantial improvement highlights the effectiveness of C$^2$Path in mitigating catastrophic forgetting under a base-category-dominant evaluation scenario. Moreover, Figure~\ref{fig:qualitative} illustrates that GCD exhibits typical failure cases, including category confusion, false positives, and missed detections of previously learned classes, which are consistent with the feature degradation observed in Figure~\ref{fig:distance_matrix}. In contrast, C$^2$Path effectively alleviates these errors while accurately recognizing novel categories.


Across both incremental settings, C$^2$Path consistently outperforms existing methods, especially on $AP_{75}$ and large-object detection, benefiting from class-conditional pathway decoupling that enables class-aware adaptation while mitigating interference from novel-category updates.

\definecolor{pottedplant_color}{HTML}{6BD0E2} \definecolor{vase_color}{HTML}{8F313F} \definecolor{person_color}{HTML}{68B259} \definecolor{snowboard_color}{HTML}{FD843F} \definecolor{surfboard_color}{HTML}{0CC9C5} \definecolor{pizza_color}{HTML}{1788CC} \definecolor{cup_color}{HTML}{693716} \definecolor{toliet_color}{HTML}{7F0104} \definecolor{sink_color}{HTML}{87A91E} \definecolor{diningtable_color}{HTML}{6BB4D4} \definecolor{bottle_color}{HTML}{BF27D4}

\begin{table}[t]
\centering
\small
\setlength{\tabcolsep}{6pt}
\begin{tabular}{l|cc|ccc}
\toprule
Baseline & CEL & CCDM& $AP$ & $AP_{50}$ & $AP_{75}$ \\ \midrule
\multirow{3}{*}{GCD} &&              & 46.7 & 63.9 & 50.8 \\
&$\surd$&     & 48.4 & 65.3 & 52.6 \\
&$\surd$& $\surd$          & \textbf{48.7} & \textbf{65.6} & \textbf{53.1} \\ \bottomrule
\end{tabular}
\caption{Ablation study on MS COCO 2017 under the 70+10 setting. CEL: Category Expert Library. CCDM: Category-Conditional Decoupling Module. ``Only CEL'' denotes that \textit{ClassLoRA} is generated by averaging all experts in CEL.}
\label{tab:ablation_study}
\end{table}

\subsection{Ablation Study}
\label{sec:ablation}
We conduct ablation studies on the two key components of C$^2$Path under the 70+10 setting, which is more sensitive to catastrophic forgetting. All variants share the same detector architecture, training strategy, distillation objectives, and parameter configuration to ensure a fair comparison. As shown in Table~\ref{tab:ablation_study}, we gradually introduce the Category Expert Library (CEL) and the Category-Conditional Decoupling Module (CCDM) to investigate their individual contributions.

\paragraph{Category Expert Library (CEL).}
Starting from the GCD baseline, introducing CEL with uniform expert aggregation improves performance from 46.7 AP to 48.4 AP. This improvement shows that maintaining a library of learnable low-rank experts provides additional adaptation capacity for novel categories while preserving the original vision-language alignment through parameter-efficient updates. Since all categories share the same expert combination, the resulting pathway remains category-agnostic and cannot explicitly mitigate class knowledge coupling.

\paragraph{Category-Conditional Decoupling Module (CCDM).}
By further introducing CCDM, C$^2$Path achieves the best performance of 48.7 AP, yielding an additional gain of 0.3 AP. The improvement is also consistent under different IoU thresholds, with a larger gain on $AP_{75}$, indicating enhanced category discrimination and localization quality. This result verifies that dynamically generating category-aware routing signals enables more effective expert selection and forms category-specific computational pathways, thereby reducing interference among categories during incremental learning.

\subsection{Analysis Experiment}
We conduct three analyses across the incremental learning phases, including Phase 1 and Phase 2. Phase 1 learns the initial representations from base categories, while Phase 2 performs exemplar-free adaptation to novel categories.

\begin{table}[t]
\centering
\small
\setlength{\tabcolsep}{6pt}
\begin{tabular}{lcccc}
\toprule
\multirow{2}{*}{Method} & Phase 1 & \multicolumn{3}{c}{Phase 2} \\
\cmidrule(lr){3-5}
 & Base & Base & Novel & All \\ \midrule
C$^2$Path (Ours) & 55.8 & 53.0 & 40.0 & 46.5 \\ \bottomrule
\end{tabular}
\caption{Performance Analysis under the 40+40 Setting. Phase 1 reports the AP of the base model on the 40 base categories, whereas Phase 2 reports the AP of the final model on the base (1--40), novel (41--80), and all (1--80) categories.}
\label{tab:stage_comparison}
\end{table}

\begin{figure}[t]
    \centering
    \begin{subfigure}[b]{0.3\columnwidth}
        \centering
        \setlength{\fboxsep}{0pt}%
        \setlength{\fboxrule}{1pt}%
        \fbox{\includegraphics[width=\textwidth]{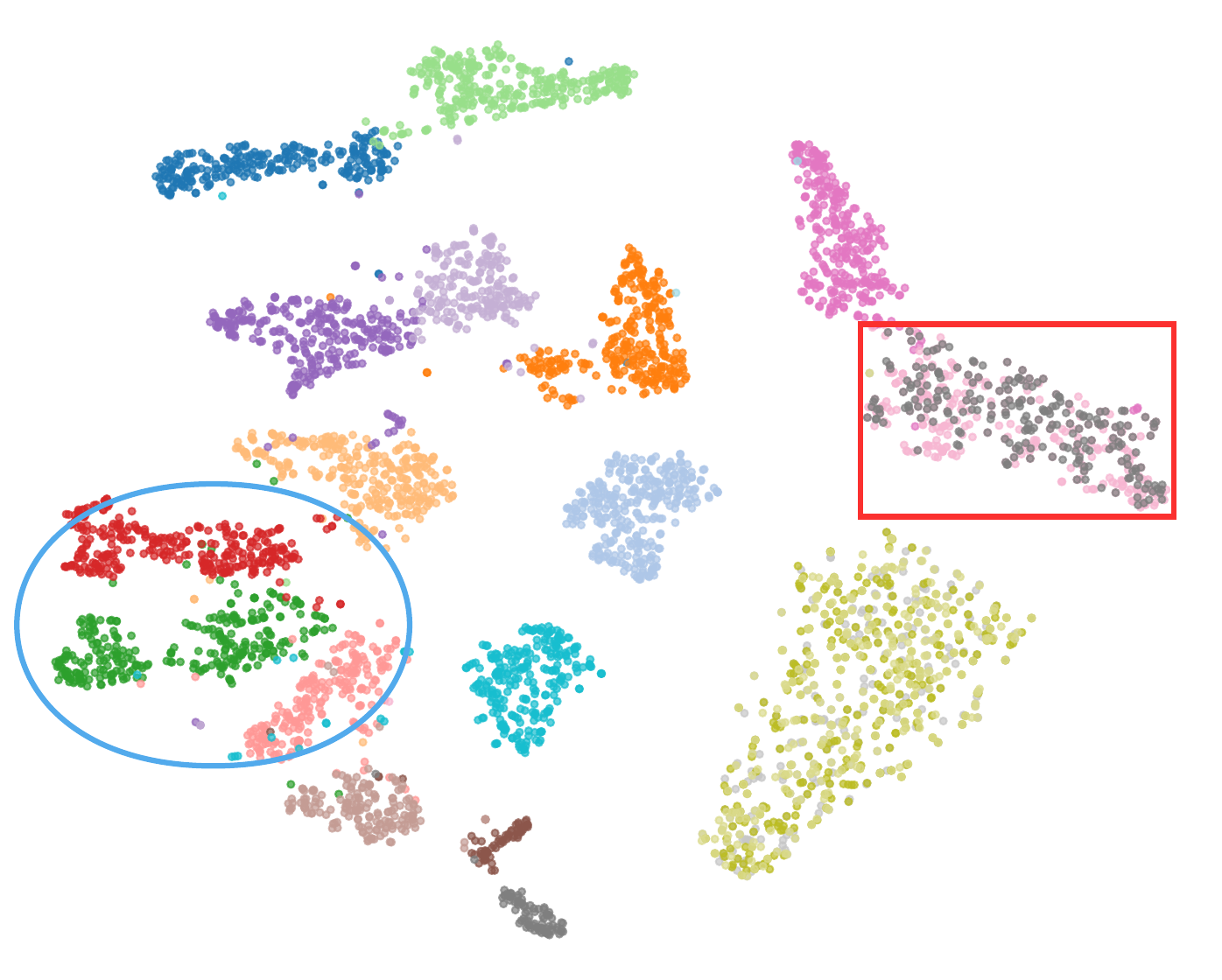}}
        \caption{Phase 1}
    \end{subfigure}
    \hfill
    \begin{subfigure}[b]{0.3\columnwidth}
        \centering
        \setlength{\fboxsep}{0pt}%
        \setlength{\fboxrule}{1pt}%
        \fbox{\includegraphics[width=\textwidth]{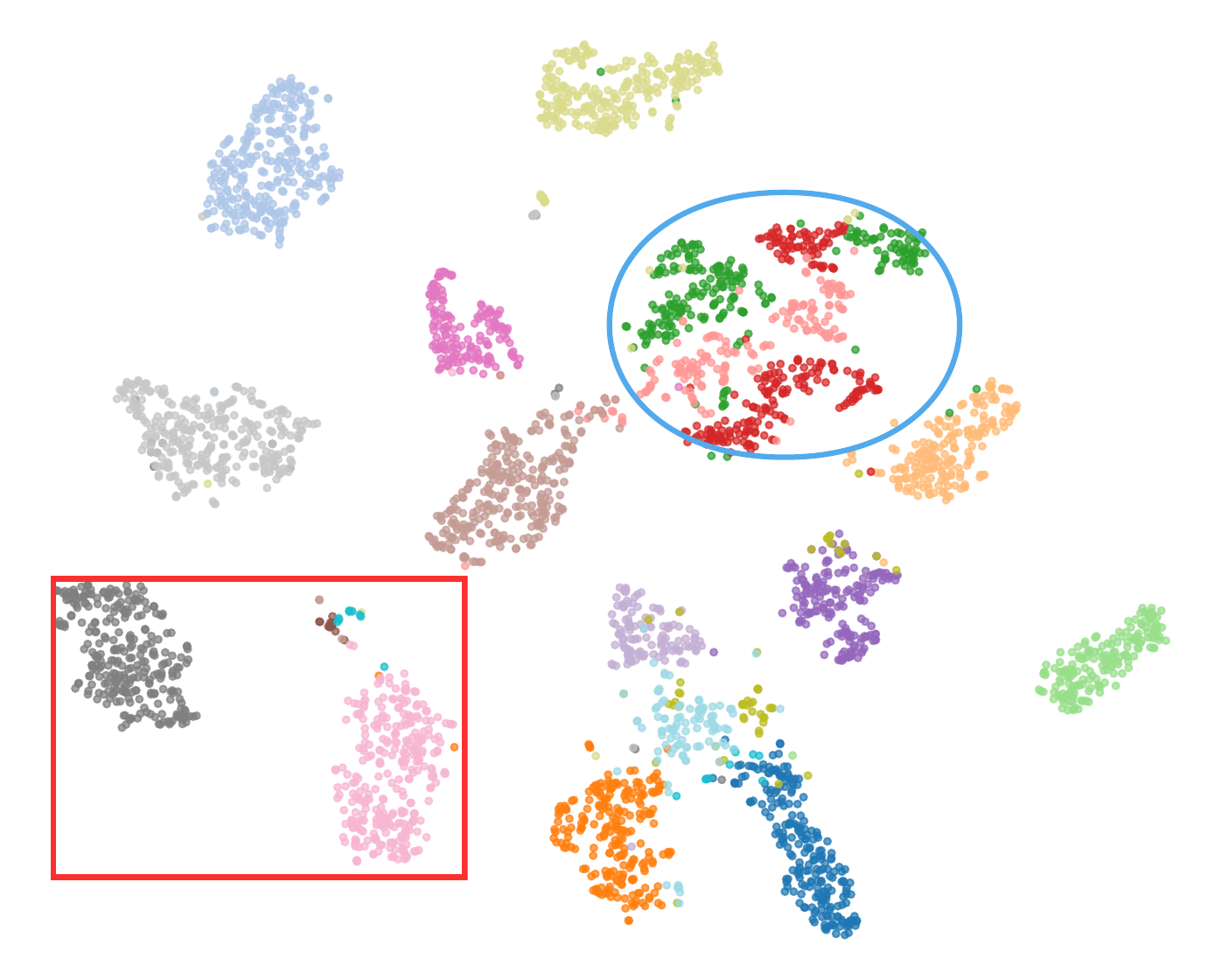}}
        \caption{Phase 2 (GCD)}
    \end{subfigure}
    \hfill
    \begin{subfigure}[b]{0.3\columnwidth}
        \centering
        \setlength{\fboxsep}{0pt}%
        \setlength{\fboxrule}{1pt}%
        \fbox{\includegraphics[width=\textwidth]{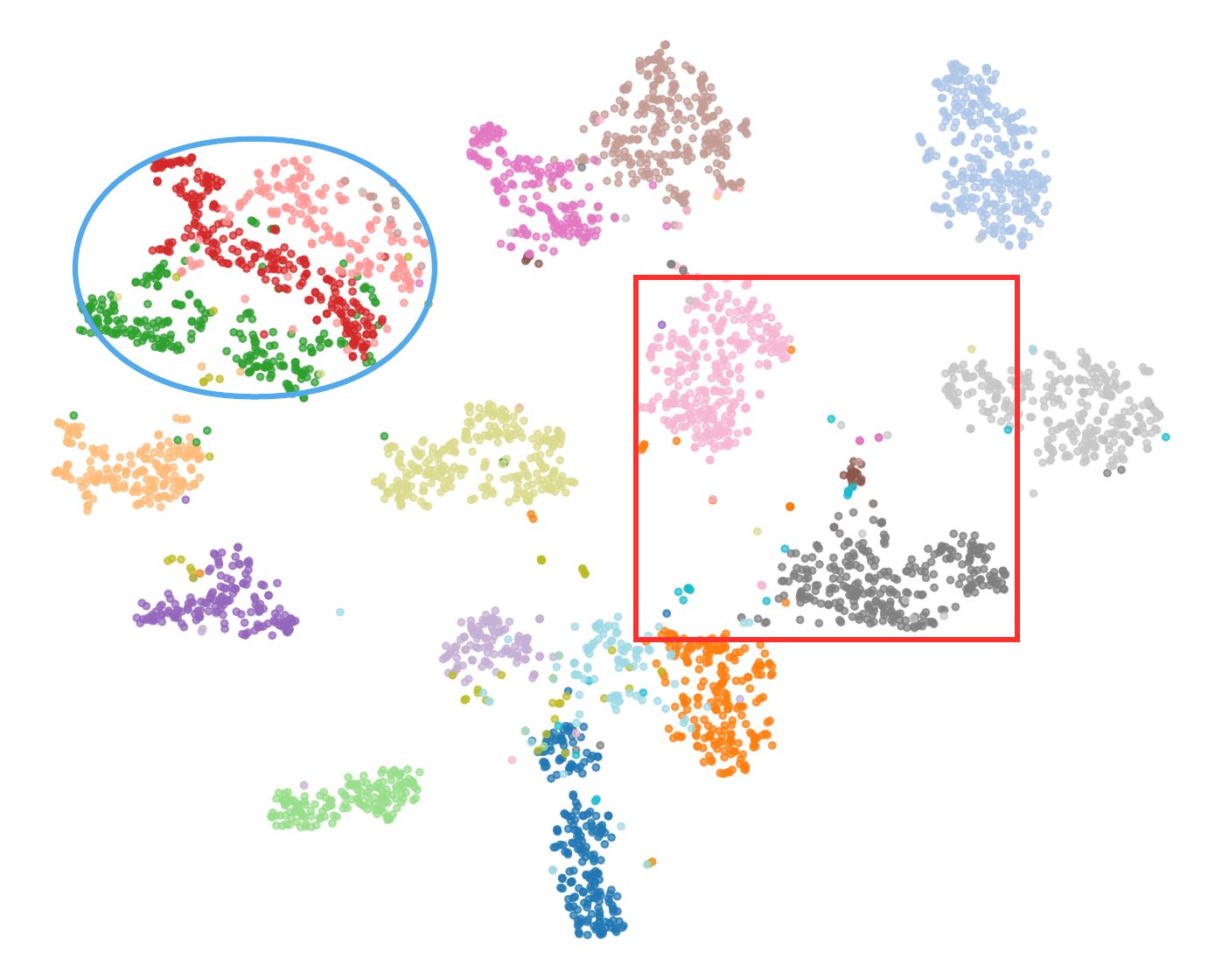}}
        \caption{Phase 2 (C$^2$Path)}
    \end{subfigure}
    \caption{Visualization of visual--linguistic features for 20 representative categories (10 base and 10 novel) after incremental training. Phase 1 learns initial representations from 70 base categories, while Phase 2 performs exemplar-free adaptation to 10 novel categories. Blue ellipses and red rectangles denote base and novel categories, respectively.}
    \label{fig:tsne}
\end{figure}

\paragraph{Base and Novel Category Performance Analysis.}
As reported in Table~\ref{tab:stage_comparison},
after adapting to novel categories, the base-category AP decreases by only 2.8 points (55.8$\rightarrow$53.0), indicating that C$^2$Path preserves approximately $95\%$ of its Phase 1 performance without accessing previous data. Meanwhile, it achieves 40.0 AP on novel categories, demonstrating an effective balance between base-category retention and novel-category acquisition.

\paragraph{Visual–Linguistic Feature Analysis.}
Figure~\ref{fig:tsne} shows the decoder output features of object queries corresponding to 20 representative categories (10 base and 10 novel) after 70+10 incremental training using t-SNE. For novel categories (red rectangle), both GCD and C$^2$Path produce well-separated clusters, indicating that both methods retain the ability to acquire novel-category knowledge. In contrast, the difference is more pronounced for base categories (blue ellipse): GCD suffers from representation drift that compromises category boundaries, whereas C$^2$Path maintains the original category-wise feature separability of the base-stage model. These results demonstrate that category-conditional pathway decoupling effectively preserves category-specific representations by enabling more isolated parameter adaptation.

\begin{figure}[t]
    \centering
    \begin{subfigure}[b]{0.3\columnwidth}
        \centering
        \includegraphics[width=\textwidth]{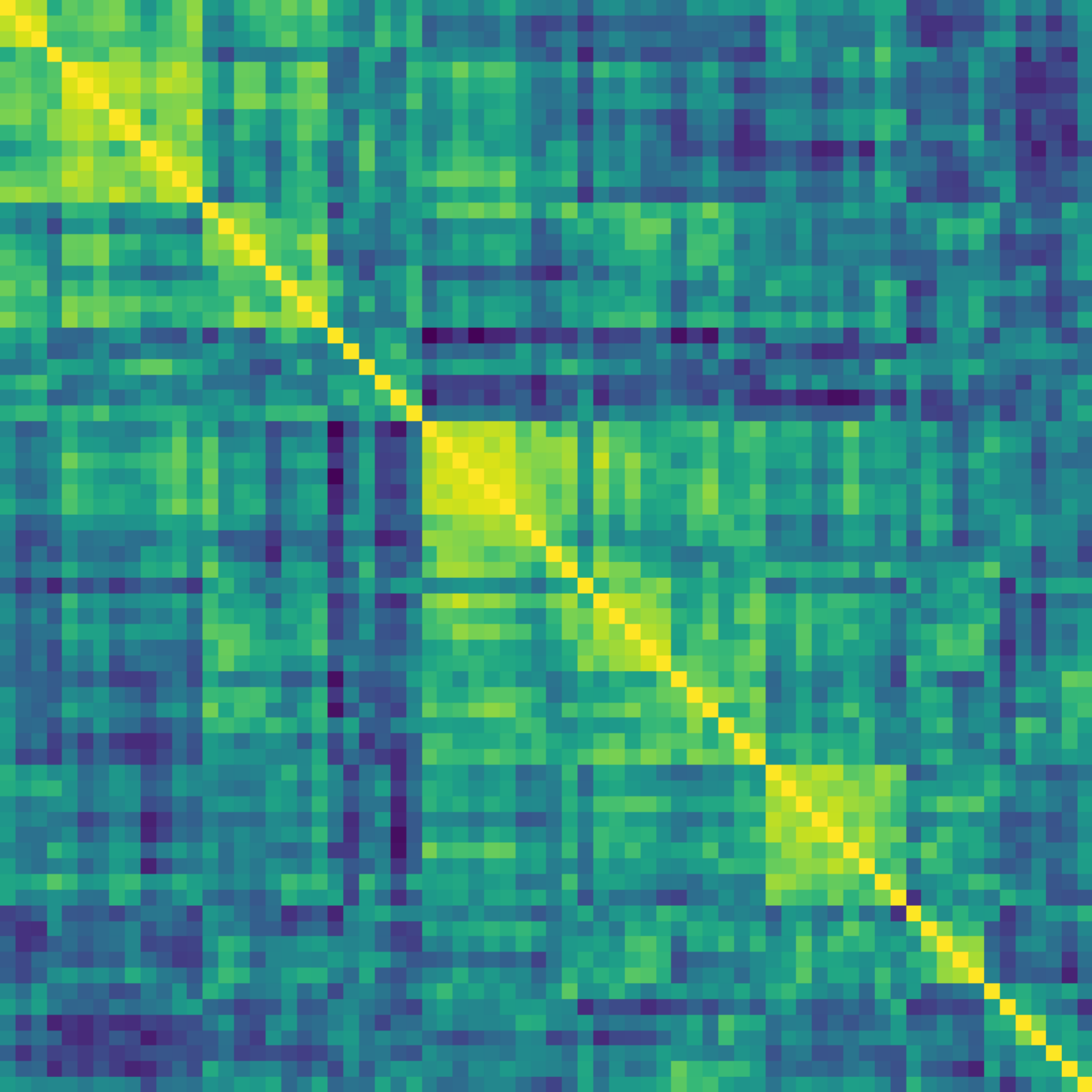}
        \caption{Phase 1}
    \end{subfigure}
    \hfill
    \begin{subfigure}[b]{0.3\columnwidth}
        \centering
        \includegraphics[width=\textwidth]{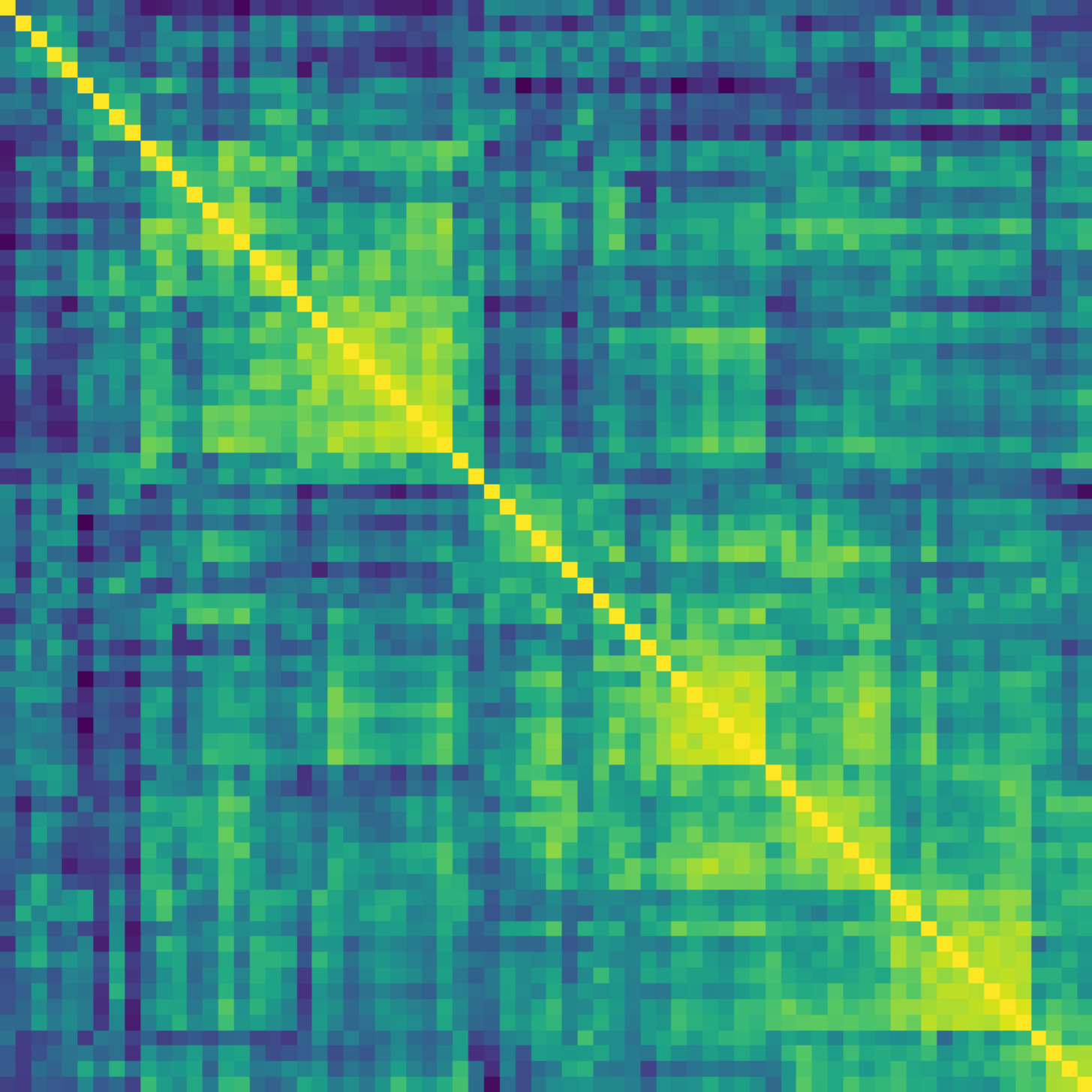}
        \caption{Phase 2 (GCD)}
    \end{subfigure}
    \hfill
    \begin{subfigure}[b]{0.3\columnwidth}
        \centering
        \includegraphics[width=\textwidth]{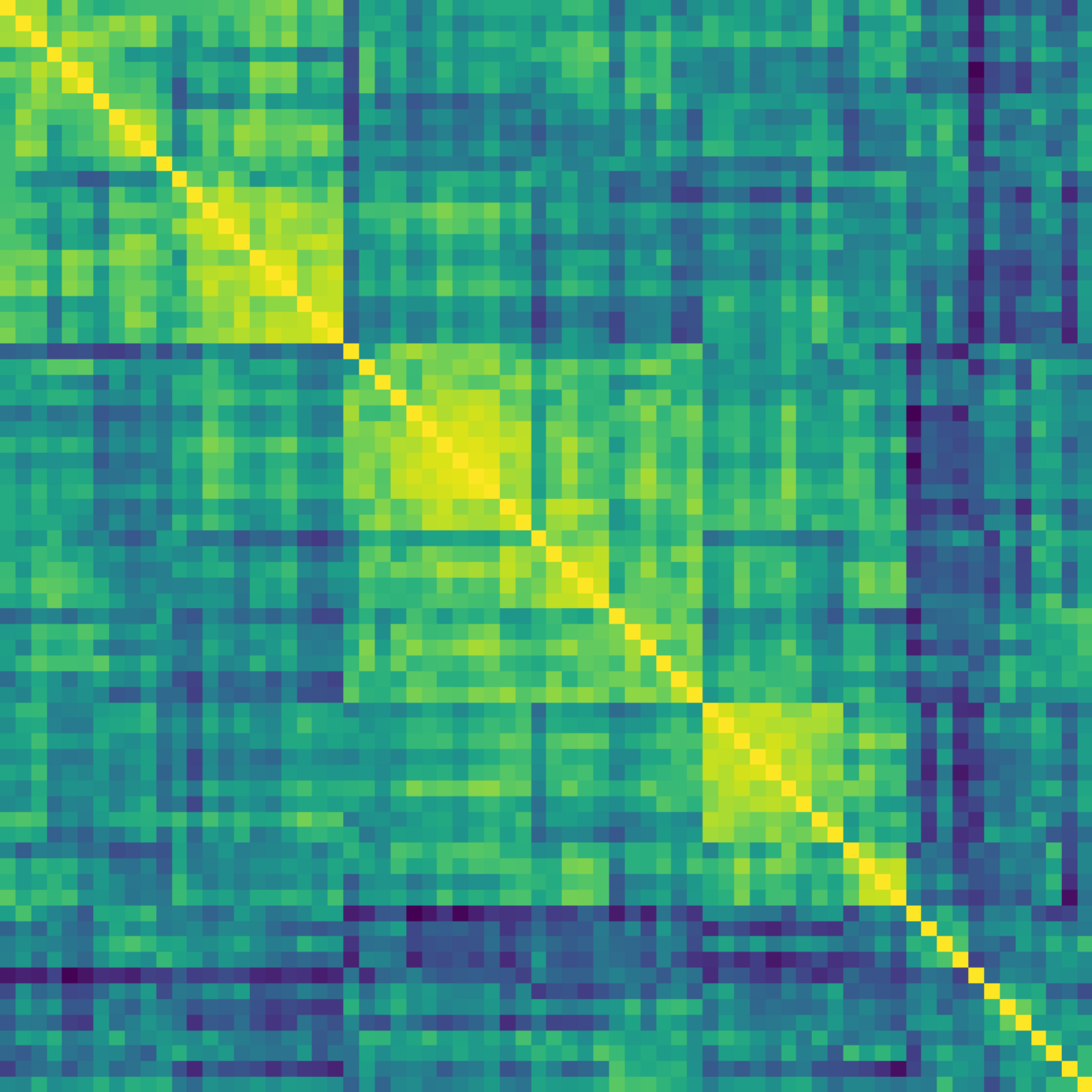}
        \caption{Phase 2 (C$^2$Path)}
    \end{subfigure}
\caption{Pairwise cosine distance matrices of the encoder-fused text features of the 70 base categories. Phase 1 learns initial representations from 70 base categories, while Phase 2 performs exemplar-free adaptation to 10 novel categories.
}
\label{fig:distance_matrix}
\end{figure}

\paragraph{Pairwise Cosine Distance Analysis of Base Categories.}
Figure~\ref{fig:distance_matrix} compares the pairwise cosine distance matrices of the encoder-fused text features for the 70 base categories before and after incremental adaptation. The Phase 1 matrix exhibits a clear and structured similarity pattern, reflecting the semantic relationships among base categories established during the initial training stage. After adapting to novel categories, GCD introduces noticeable distortions into the original feature geometry: the global pattern becomes less consistent, and the contrast between semantically related and unrelated categories is weakened, indicating that incremental updates interfere with the previously learned text representations. In contrast, C$^2$Path largely preserves the structural characteristics of the Phase 1 matrix, maintaining both the overall distribution pattern and the relative distances among base-category embeddings. This demonstrates that category-conditional pathway decoupling effectively isolates novel-category updates from the established base-category representations, thereby preventing semantic drift in the text feature space and maintaining the stability of the original vision-language alignment.


\section{Conclusion}
In this work, we reveal a \textbf{class knowledge coupling} issue in existing vision-language incremental object detection. We argue that continual category expansion requires class-conditional pathway learning to enable independent and composable knowledge evolution. To this end, we propose \textbf{C$^2$Path}, a pathway decoupling framework that transforms global parameter adaptation into semantic-guided composition of class-specific computational pathways through dynamically synthesized \textit{ClassLoRA} adapters. Extensive experiments demonstrate that C$^2$Path consistently outperforms existing methods, validating the effectiveness of class-conditional computational pathway decoupling for scalable vision-language incremental object detection.




{
    \small
    \bibliographystyle{ieeenat_fullname}
    \bibliography{refs}

@article{peng2020faster,
  title={Faster ilod: Incremental learning for object detectors based on faster rcnn},
  author={Peng, Can and Zhao, Kun and Lovell, Brian C},
  journal={Pattern recognition letters},
  volume={140},
  pages={109--115},
  year={2020},
  publisher={Elsevier}
}

@inproceedings{shmelkov2017incremental,
  title={Incremental learning of object detectors without catastrophic forgetting},
  author={Shmelkov, Konstantin and Schmid, Cordelia and Alahari, Karteek},
  booktitle={Proceedings of the IEEE international conference on computer vision},
  pages={3400--3409},
  year={2017}
}

@article{peng2021sid,
  title={Sid: Incremental learning for anchor-free object detection via selective and inter-related distillation},
  author={Peng, Can and Zhao, Kun and Maksoud, Sam and Li, Meng and Lovell, Brian C},
  journal={Computer vision and image understanding},
  volume={210},
  pages={103229},
  year={2021},
  publisher={Elsevier}
}

@inproceedings{joseph2021towards,
  title={Towards open world object detection},
  author={Joseph, KJ and Khan, Salman and Khan, Fahad Shahbaz and Balasubramanian, Vineeth N},
  booktitle={Proceedings of the IEEE/CVF conference on computer vision and pattern recognition},
  pages={5830--5840},
  year={2021}
}

@article{joseph2021incremental,
  title={Incremental object detection via meta-learning},
  author={Joseph, KJ and Rajasegaran, Jathushan and Khan, Salman and Khan, Fahad Shahbaz and Balasubramanian, Vineeth N},
  journal={IEEE Transactions on Pattern Analysis and Machine Intelligence},
  volume={44},
  number={12},
  pages={9209--9216},
  year={2021},
  publisher={IEEE}
}

@inproceedings{gupta2022ow,
  title={Ow-detr: Open-world detection transformer},
  author={Gupta, Akshita and Narayan, Sanath and Joseph, KJ and Khan, Salman and Khan, Fahad Shahbaz and Shah, Mubarak},
  booktitle={Proceedings of the IEEE/CVF conference on computer vision and pattern recognition},
  pages={9235--9244},
  year={2022}
}

@inproceedings{liu2023continual,
  title={Continual detection transformer for incremental object detection},
  author={Liu, Yaoyao and Schiele, Bernt and Vedaldi, Andrea and Rupprecht, Christian},
  booktitle={Proceedings of the IEEE/CVF conference on computer vision and pattern recognition},
  pages={23799--23808},
  year={2023}
}

@article{deng2024zero,
  title={Zero-shot generalizable incremental learning for vision-language object detection},
  author={Deng, Jieren and Zhang, Haojian and Ding, Kun and Hu, Jianhua and Zhang, Xingxuan and Wang, Yunkuan},
  journal={Advances in Neural Information Processing Systems},
  volume={37},
  pages={136679--136700},
  year={2024}
}

@inproceedings{kim2024vlm,
  title={Vlm-pl: Advanced pseudo labeling approach for class incremental object detection via vision-language model},
  author={Kim, Junsu and Ku, Yunhoe and Kim, Jihyeon and Cha, Junuk and Baek, Seungryul},
  booktitle={Proceedings of the IEEE/CVF Conference on Computer Vision and Pattern Recognition},
  pages={4170--4181},
  year={2024}
}

@inproceedings{wang2025gcd,
  title={Gcd: Advancing vision-language models for incremental object detection via global alignment and correspondence distillation},
  author={Wang, Xu and Wang, Zilei and Lin, Zihan},
  booktitle={Proceedings of the AAAI Conference on Artificial Intelligence},
  volume={39},
  pages={8015--8023},
  year={2025}
}

@inproceedings{yu2024boosting,
  title={Boosting continual learning of vision-language models via mixture-of-experts adapters},
  author={Yu, Jiazuo and Zhuge, Yunzhi and Zhang, Lu and Hu, Ping and Wang, Dong and Lu, Huchuan and He, You},
  booktitle={Proceedings of the IEEE/CVF Conference on Computer Vision and Pattern Recognition},
  pages={23219--23230},
  year={2024}
}

@inproceedings{he2025cl,
  title={CL-LoRA: Continual low-rank adaptation for rehearsal-free class-incremental learning},
  author={He, Jiangpeng and Duan, Zhihao and Zhu, Fengqing},
  booktitle={Proceedings of the Computer Vision and Pattern Recognition Conference},
  pages={30534--30544},
  year={2025}
}

@inproceedings{wang2025self,
  title={Self-expansion of pre-trained models with mixture of adapters for continual learning},
  author={Wang, Huiyi and Lu, Haodong and Yao, Lina and Gong, Dong},
  booktitle={Proceedings of the Computer Vision and Pattern Recognition Conference},
  pages={10087--10098},
  year={2025}
}

@article{wu2025sd,
  title={Sd-lora: Scalable decoupled low-rank adaptation for class incremental learning},
  author={Wu, Yichen and Piao, Hongming and Huang, Long-Kai and Wang, Renzhen and Li, Wanhua and Pfister, Hanspeter and Meng, Deyu and Ma, Kede and Wei, Ying},
  journal={arXiv preprint arXiv:2501.13198},
  year={2025}
}

@article{dong2024mr,
  title={MR-GDINO: efficient open-world continual object detection},
  author={Dong, Bowen and Huang, Zitong and Yang, Guanglei and Zhang, Lei and Zuo, Wangmeng},
  journal={arXiv preprint arXiv:2412.15979},
  year={2024}
}

@inproceedings{zhang2024learning,
  title={Learning task-aware language-image representation for class-incremental object detection},
  author={Zhang, Hongquan and Gao, Bin-Bin and Zeng, Yi and Tian, Xudong and Tan, Xin and Zhang, Zhizhong and Qu, Yanyun and Liu, Jun and Xie, Yuan},
  booktitle={Proceedings of the AAAI Conference on Artificial Intelligence},
  volume={38},
  pages={7096--7104},
  year={2024}
}

@inproceedings{liu2024grounding,
  title={Grounding dino: Marrying dino with grounded pre-training for open-set object detection},
  author={Liu, Shilong and Zeng, Zhaoyang and Ren, Tianhe and Li, Feng and Zhang, Hao and Yang, Jie and Jiang, Qing and Li, Chunyuan and Yang, Jianwei and Su, Hang and others},
  booktitle={European conference on computer vision},
  pages={38--55},
  year={2024},
  organization={Springer}
}

@inproceedings{kang2023alleviating,
  title={Alleviating catastrophic forgetting of incremental object detection via within-class and between-class knowledge distillation},
  author={Kang, Mengxue and Zhang, Jinpeng and Zhang, Jinming and Wang, Xiashuang and Chen, Yang and Ma, Zhe and Huang, Xuhui},
  booktitle={Proceedings of the IEEE/CVF International Conference on Computer Vision},
  pages={18894--18904},
  year={2023}
}

@article{hu2022lora,
  title={Lora: Low-rank adaptation of large language models.},
  author={Hu, Edward J and Shen, Yelong and Wallis, Phillip and Allen-Zhu, Zeyuan and Li, Yuanzhi and Wang, Shean and Wang, Liang and Chen, Weizhu and others},
  journal={Iclr},
  volume={1},
  number={2},
  pages={3},
  year={2022}
}

@article{li2017learning,
  title={Learning without forgetting},
  author={Li, Zhizhong and Hoiem, Derek},
  journal={IEEE transactions on pattern analysis and machine intelligence},
  volume={40},
  number={12},
  pages={2935--2947},
  year={2017},
  publisher={IEEE}
}

@inproceedings{feng2022overcoming,
  title={Overcoming catastrophic forgetting in incremental object detection via elastic response distillation},
  author={Feng, Tao and Wang, Mang and Yuan, Hangjie},
  booktitle={Proceedings of the IEEE/CVF conference on computer vision and pattern recognition},
  pages={9427--9436},
  year={2022}
}

@inproceedings{kim2024sddgr,
  title={Sddgr: Stable diffusion-based deep generative replay for class incremental object detection},
  author={Kim, Junsu and Cho, Hoseong and Kim, Jihyeon and Tiruneh, Yihalem Yimolal and Baek, Seungryul},
  booktitle={Proceedings of the IEEE/CVF Conference on Computer Vision and Pattern Recognition},
  pages={28772--28781},
  year={2024}
}

@article{kirkpatrick2017overcoming,
  title={Overcoming catastrophic forgetting in neural networks},
  author={Kirkpatrick, James and Pascanu, Razvan and Rabinowitz, Neil and Veness, Joel and Desjardins, Guillaume and Rusu, Andrei A and Milan, Kieran and Quan, John and Ramalho, Tiago and Grabska-Barwinska, Agnieszka and others},
  journal={Proceedings of the national academy of sciences},
  volume={114},
  number={13},
  pages={3521--3526},
  year={2017},
  publisher={National Academy of Sciences}
}

@inproceedings{radford2021learning,
  title={Learning transferable visual models from natural language supervision},
  author={Radford, Alec and Kim, Jong Wook and Hallacy, Chris and Ramesh, Aditya and Goh, Gabriel and Agarwal, Sandhini and Sastry, Girish and Askell, Amanda and Mishkin, Pamela and Clark, Jack and others},
  booktitle={International conference on machine learning},
  pages={8748--8763},
  year={2021},
  organization={PmLR}
}

@article{von2019continual,
  title={Continual learning with hypernetworks},
  author={Von Oswald, Johannes and Henning, Christian and Grewe, Benjamin F and Sacramento, Jo{\~a}o},
  journal={arXiv preprint arXiv:1906.00695},
  year={2019}
}

@inproceedings{carion2020end,
  title={End-to-end object detection with transformers},
  author={Carion, Nicolas and Massa, Francisco and Synnaeve, Gabriel and Usunier, Nicolas and Kirillov, Alexander and Zagoruyko, Sergey},
  booktitle={European conference on computer vision},
  pages={213--229},
  year={2020},
  organization={Springer}
}

@inproceedings{lin2014microsoft,
  title={Microsoft coco: Common objects in context},
  author={Lin, Tsung-Yi and Maire, Michael and Belongie, Serge and Hays, James and Perona, Pietro and Ramanan, Deva and Doll{\'a}r, Piotr and Zitnick, C Lawrence},
  booktitle={European conference on computer vision},
  pages={740--755},
  year={2014},
  organization={Springer}
}

@inproceedings{li2022glip,
  title={Grounded Language-Image Pre-training},
  author={Li, Liunian Harold and Zhang, Pengchuan and Zhang, Haotian and
          Yang, Jianwei and Li, Chunyuan and Zhong, Yiwu and Wang, Lijuan and
          Yuan, Lu and Zhang, Lei and Hwang, Jenq-Neng and Chang, Kai-Wei and
          Gao, Jianfeng},
  booktitle={Proceedings of the IEEE/CVF Conference on Computer Vision and
             Pattern Recognition (CVPR)},
  pages={10965--10975},
  year={2022}
}

@inproceedings{ha2017hypernetworks,
  title={HyperNetworks},
  author={Ha, David and Dai, Andrew M. and Le, Quoc V.},
  booktitle={International Conference on Learning Representations (ICLR)},
  year={2017}
}

@inproceedings{wang2022l2p,
  title={Learning to Prompt for Continual Learning},
  author={Wang, Zifeng and Zhang, Zizhao and Lee, Chen-Yu and Zhang, Han and
          Sun, Ruoxi and Ren, Xiaoqi and Su, Guolong and Perot, Vincent and
          Dy, Jennifer and Pfister, Tomas},
  booktitle={Proceedings of the IEEE/CVF Conference on Computer Vision and
             Pattern Recognition (CVPR)},
  pages={139--149},
  year={2022}
}

@inproceedings{smith2023coda,
  title={{CODA-Prompt}: COntinual Decomposed Attention-Based Prompting for
         Rehearsal-Free Continual Learning},
  author={Smith, James Seale and Karlinsky, Leonid and Gutta, Vyshnavi and
          Cascante-Bonilla, Paola and Kim, Donghyun and Arbelle, Assaf and
          Panda, Rameswar and Feris, Rogerio and Kira, Zsolt},
  booktitle={Proceedings of the IEEE/CVF Conference on Computer Vision and
             Pattern Recognition (CVPR)},
  pages={11909--11919},
  year={2023}
}

@inproceedings{wang2023olora,
  title={Orthogonal Subspace Learning for Language Model Continual Learning},
  author={Wang, Xiao and Chen, Tianze and Ge, Qiming and Xia, Han and
          Bao, Rong and Zheng, Rui and Zhang, Qi and Gui, Tao and
          Huang, Xuanjing},
  booktitle={Findings of the Association for Computational Linguistics:
             EMNLP 2023},
  pages={10658--10671},
  year={2023}
}

@inproceedings{liang2024inflora,
  title={{InfLoRA}: Interference-Free Low-Rank Adaptation for Continual
         Learning},
  author={Liang, Yan-Shuo and Li, Wu-Jun},
  booktitle={Proceedings of the IEEE/CVF Conference on Computer Vision and
             Pattern Recognition (CVPR)},
  pages={23638--23647},
  year={2024}
}

@article{zou2023object,
  title={Object detection in 20 years: A survey},
  author={Zou, Zhengxia and Chen, Keyan and Shi, Zhenwei and Guo, Yuhong and Ye, Jieping},
  journal={Proceedings of the IEEE},
  volume={111},
  number={3},
  pages={257--276},
  year={2023},
  publisher={IEEE}
}

@article{zhao2019object,
  title={Object detection with deep learning: A review},
  author={Zhao, Zhong-Qiu and Zheng, Peng and Xu, Shou-tao and Wu, Xindong},
  journal={IEEE transactions on neural networks and learning systems},
  volume={30},
  number={11},
  pages={3212--3232},
  year={2019},
  publisher={IEEE}
}

@article{parisi2018continual,
  title={Continual lifelong learning with neural networks: A review},
  author={Parisi, German I and Kemker, Ronald and Part, Jose L and Kanan, Christopher and Wermter, Stefan},
  journal={arXiv preprint arXiv:1802.07569},
  volume={990},
  year={2018},
  publisher={eprint}
}

@inproceedings{kemker2018measuring,
  title={Measuring catastrophic forgetting in neural networks},
  author={Kemker, Ronald and McClure, Marc and Abitino, Angelina and Hayes, Tyler and Kanan, Christopher},
  booktitle={Proceedings of the AAAI conference on artificial intelligence},
  volume={32},
  number={1},
  year={2018}
}
}

\end{document}